\documentclass[letterpaper]{article} 
\usepackage{aaai24}  
\usepackage{times}  
\usepackage{helvet}  
\usepackage{courier}  
\usepackage[hyphens]{url}  
\usepackage{graphicx} 
\usepackage{natbib}  
\usepackage{caption} 
\usepackage{algorithm}
\usepackage{algorithmic}
\usepackage{amsmath}
\usepackage{newfloat}
\usepackage{listings}
\DeclareCaptionStyle{ruled}{labelfont=normalfont,labelsep=colon,strut=off} 
\floatstyle{ruled}
\newfloat{listing}{tb}{lst}{}
\floatname{listing}{Listing}
\title{QLABGrad: a Hyperparameter-Free and \\Convergence-Guaranteed Scheme for Deep Learning}
\author{
    Minghan Fu\textsuperscript{\rm 1}, Fang-Xiang Wu\textsuperscript{\rm 1,2}
}
\affiliations{
    \textsuperscript{\rm 1}Department of Mechanical Engineering, University of Saskatchewan\\
    \textsuperscript{\rm 2}Department of Computer Science, University of Saskatchewan\\
    \{fig072, faw341\}@mail.usask.ca
}

\usepackage{bibentry}

\begin{document}

\maketitle

\begin{abstract}
The learning rate is a critical hyperparameter for deep learning tasks since it determines the extent to which the model parameters are updated during the learning course. However, the choice of learning rates typically depends on empirical judgment, which may not result in satisfactory outcomes without intensive try-and-error experiments. In this study, we propose a novel learning rate adaptation scheme called QLABGrad. Without any user-specified hyperparameter, QLABGrad automatically determines the learning rate by optimizing the Quadratic Loss Approximation-Based (QLAB) function for a given gradient descent direction, where only one extra forward propagation is required.  We theoretically prove the convergence of QLABGrad with a smooth Lipschitz condition on the loss function. Experiment results on multiple architectures, including MLP, CNN, and ResNet, on MNIST, CIFAR10, and ImageNet datasets,  demonstrate that QLABGrad outperforms various competing schemes for deep learning. 
\end{abstract}

\section{Introduction}\label{introduction}
The most common scheme for learning parameters of deep network models is the gradient (steepest) descent method, in which the gradient of the loss function with a learning rate is used to update the parameters in an iterative course. The basic descent methods  include stochastic gradient descent (SGD), mini-batch gradient descent and batch gradient descent \cite{Aggarwal2018, Goodfellow2016, GDoverview} and their variants \cite{GDvariant1,GDvariant2}, where a constant finite learning rate is user-specified or is tuned with less or no adaptation. In principle,  the gradient descent direction is the optimal direction with infinitesimal learning rates. However, if a finite learning rate is chosen to be too small, the learning course could be very time-consuming. On the other hand, if a finite learning rate is chosen to be too large, the learning course can behave unexpectedly (e.g., overshooting or diverging). Therefore, in either case, maintaining a constant learning rate is not an ideal learning strategy. 

\begin{table*}[h]
\centering
\begin{tabular}{lccccr}
\hline\\
Scheme & $\Delta \theta_t$&Adaptive rule & HyperP \\
SGD& $-\alpha\nabla_\theta \mathcal{L}(\theta_{t-1})$ &-& $\alpha$\\
\hline\\
E-decay & $-\alpha f(t)\nabla_\theta \mathcal{L}(\theta_{t-1})$ &$f(t)=\exp(-\beta t)$ & $\alpha, \beta $ \\
R-decay & $-\alpha f(t)\nabla_\theta \mathcal{L}(\theta_{t-1})$ & $f(t)=1/(1+\beta t)$ &$\alpha, \beta$ \\
SS-decay & $-\alpha f(t)\nabla_\theta \mathcal{L}(\theta_{t-1})$ &$f(t)=2^{-\lfloor t/T\rfloor} $ & $\alpha, T$\\
CA-decay & $-\alpha f(t)\nabla_\theta \mathcal{L}(\theta_{t-1})$ & $f(t)=\alpha+\frac{1}{2}\left(\beta-\alpha\right)\left(1+\cos \left(\frac{T} \pi\right)\right)$& $\alpha$, $\beta, T$ \\
\hline\\
Adagrad& $-\frac{\alpha}{\sqrt{A_t}}\nabla_\theta\mathcal{L}(\theta_{t-1})$ & $A_t= A_{t-1}+\|\nabla_\theta \mathcal{L}(\theta_{t-1})\|^2$ &$\alpha$\\
RMSprop& $-\frac{\alpha}{\sqrt{A_t}}\nabla_\theta \mathcal{L}(\theta_{t-1})$ & $A_t= \beta A_{t-1}+(1-\beta)\|\nabla_\theta \mathcal{L}(\theta_{t-1})\|^2$ &$ \alpha, \beta$\\
Adadelta & $-\sqrt{\frac{B_t}{A_t}}\nabla_\theta \mathcal{L}(\theta_{t-1})$ & $\begin{array}{l} A_t=\beta A_{t-1}+(1-\beta)\|\nabla_\theta \mathcal{L}(\theta_{t-1})\|^2 \\ B_t= \beta B_{t-1}+(1-\beta)\|\Delta\theta_{t-1}\|^2 \end{array}$ & $\beta$\\
HGD & $-A_t \nabla_\theta\mathcal{L}(\theta_{t-1})$ & $A_t= A_{t-1}+\beta \nabla_\theta \mathcal{L} (\theta_{t-2})\nabla_\theta \mathcal{L}(\theta_{t-1})$ & $\beta$\\
L4GD & $-\alpha A_t \nabla_\theta\mathcal{L}(\theta_{t-1})$ & $A_t =\frac{\mathcal{L}(\theta_{t-1})-\beta \mathcal{L}^{min}}{\|\nabla_\theta \mathcal{L}(\theta_{t-1})\|^2} $ & $\alpha, \beta$\\
LQA & $-A_t \nabla_\theta\mathcal{L}(\theta_{t-1})$ & $A_t=\frac{1}{2} \alpha+\frac{L\left(\theta_{t-1}+\alpha \nabla L\left(\theta_{t-1}\right)\right)-L\left(\theta_{t-1}-\alpha \nabla L\left(\theta_{t-1}\right)\right)}{L\left(\theta_{t-1}+\alpha \nabla L\left(\theta_{t-1}\right)\right)+L\left(\theta_{t-1}-\alpha \nabla L\left(\theta_{t-1}\right)\right)-2 L\left(\theta_{t-1}\right)}$ & $\alpha$\\
\hline\\
Momentum& $-\alpha F_t$ & $F_t= \gamma F_{t-1}+(1-\gamma)\nabla_\theta \mathcal{L}(\theta_{t-1})$ & $\alpha, \gamma$\\
NMomentum & $-\alpha F_t$ & $\begin{array}{l} F_t= \gamma F_t-(1-\gamma)\nabla_\theta \mathcal{L}(\theta_{t-1}+\gamma \Delta \theta_{t-1}) \end{array}$ & $\alpha, \gamma$ \\
Adam & $-\alpha\left(\frac{\sqrt{1-\beta^t}}{1-\gamma^t}\right)\frac{F_t}{\sqrt{A_t}}$ & $\begin{array}{l} A_t= \beta A_{t-1}+(1-\beta)\|\nabla_\theta \mathcal{L}(\theta_{t-1})\|^2 \\ F_t= \gamma F_{t-1} + (1-\gamma)\nabla_\theta \mathcal{L}(\theta_{t-1}) \end{array}$ & $\alpha, \beta, \gamma$ \\
\hline\\
\end{tabular}
\caption{The most common schemes for deep learning.}
\label{Tablem}
\end{table*}

Several strategies have been proposed to address the issues with a constant learning rate as summarized in Table \ref{Tablem}. The first strategy is the learning rate decay with the iteration number $t$, where the learning rate is expressed as the product of the hyperparameter $\alpha$ and a monotonically decreasing function of  $t$, while the descent direction is the gradient. There are several commonly used monotonically decreasing functions for adapting the learning rate for deep learning tasks, including exponential decay (E-decay) , reciprocal decay (R-decay), step-size decay (SS-decay), and cosine annealing decay (CA-decay) schemes. 

Another strategy makes use of the loss function value and/or the loss function gradient for adapting the learning rate while the decent direction is also the gradient. AdaGrad \cite{adagrad} is the first method in this category, which aggregates squared gradient norms over the learning course, denoted as $A_t$. The square-root of $A_t$ is used to adjust the learning rate. Instead of simply aggregating the squared gradient norms, RMSprop \cite{rms} calculates $A_t$ with the exponential moving average of squared gradient norms. Furthermore, AdaDelta \cite{Zeiler2012} replaces the learning rate in RMSProp with the square root of the exponential moving average of squared incremental update norms. As a result, there is no learning rate in AdaDelta. More recently the hypergradient decent (HGD) is rediscovered to learn the learning rate \cite{Baydin2018,L2O2022}, which originally considers the adaptation of the learning rate using the derivative of the objective function with respect to the learning rate \cite{Almeida1998}. Computing this hypergradient requires one extra copy of the original gradient to be stored in memory. It has been illustrated that the hypergradient decent can be more robust by using the gradient-related information of up to two previous steps in adapting the learning rate \cite{Plagianakos2001}. L4 \cite{l4} adapts the learning rate with the current and minimum loss function values, as well as the current gradient and another descent direction. L4GD utilizes the gradient as the descent direction, while L4Mom and L4Adam \cite{l4} apply Momentum and Adam's update directions, respectively. LQA \cite{lqa} employs the three loss function values to calculate the adaptive learning rate at each iteration.

The third strategy can be viewed as the schemes with a constant learning rate while adapting the descent direction. Momentum \cite{Jacobs1988} keeps a discounted history of past gradients to build a momentum \cite{Polyak1964} in a search direction calculated with the exponetial moving average of gradients. Nesterov momentum (NMomentum) \cite{Nesterov1983} is a modification of the traditional momentum method in which the gradients are computed at a shifted point, as is common in the Nesterov method \cite{Nestrov1983}. RMSprop with Nesterov momentum (RMSprop-NM) uses the gradient of the loss function at a shifted point, to calculate $A_t$ and then use Nesterov momentum to improve the descent direction. Adam \cite{adam} adapts the descent direction by multiplying the exponential moving average of gradients with a bias function depending on the iteration index $t$ and the exponential moving average of squared gradient norms. 

As shown in the last column of Table \ref{Tablem}, all schemes need one up to three user-specified hyperparameters. The specification of these hyperparameters is challenging as an improper value can lead to a significant decrease of the performance of deep learning models.  In this study, we mainly focus on developing a hyperparameter-free and convergence-guaranteed scheme for adapting the learning rate while the decent direction is the gradient, named QLABGrad. Besides the theoretical analysis, we conduct extensive experiments to illustrate the performance of QLABGrad for deep learning, compared to those competing schemes in Table \ref{Tablem}.  

\section{Method}
\subsection{Motivation}
 The first motivation of our study can be well explained with Table \ref{Tablem}. All schemes mainly try to improve the basic gradient decent (GD) scheme which has only one hyperparameter. However, as can be seen, majority of these improved schemes introduce at least one more hyperparameters. This means that in practice users still have to choose one or more hyperparameters, for example, Adam needs users to specify three hyperparameters. Although it is wished that the introduced hyperparameters are insensitive to the learning course \cite{Baydin2018,Jie2022,adamhypers}, this may not be the case in practice. In summary, the schemes in Table \ref{Tablem} still require users to properly specify one or more parameters according to their experiences, which can be challenging \cite{hypertuning,hypertuning2}. 
 
Our second motivation comes from the derivation of L4 by Rolinek and Martius \cite{l4}.  L4 is derived from the linear approximation of the loss function to estimate the proper learning rate \cite{BeckandArnold1977}. To avoid the overestimation of the learning rate, a couple of extra hyperparameters are introduced. Besides the specification of these hyperparameters, the estimation of the minimum loss function could be problematic in practice. 

Our third motivation arises from considering the limitations of LQA \cite{lqa}. LQA determines the adaptive learning rate by requiring two additional loss function values at each iteration, which could introduce a significant computational burden during the learning course. In addition, LQA does not guarantee the convergence.

In this study, we propose a quadratic loss approximation-based (QLAB) method to automatically determine the optimal learning rate without the requirement of any wisely user-specified hyperparameter when the descent direction is the gradient. The idea of using the quadratic loss approximation to determine the learning rate (similar to LQA\cite{lqa}) can back to the Box-Kanemasu’s method \cite{BeckandArnold1977}, which tries to address the overshooting problem in the traditional Newton's method.

\begin{figure}[t]
\centering
\includegraphics[width=0.49\textwidth]{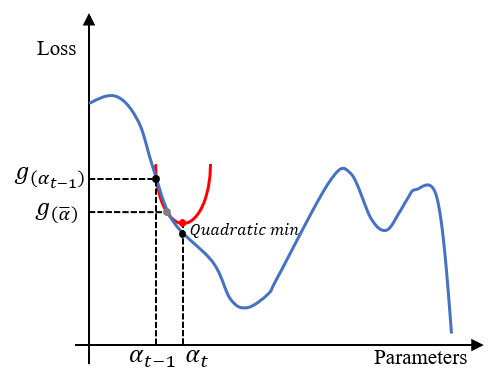}
\caption{Illustration of QLAB.}
\label{illustation}
\end{figure}

\subsection{Derivation of QLAB Algorithm} \label{DQlab}
Let $\mathcal{L}(\theta)$ be the loss function depending on model parameters $\theta\in\mathbf{R}^d$, we are interested in finding the minimizer:
\begin{equation}
\theta^*=arg\min_{\theta\in\mathbf{R}^d} \mathcal{L}(\theta)
\end{equation}
where $d$ is the number of model parameters.

Given the parameter values $\theta_{t-1}$ at the $t-1$ iterative step, the gradient decent scheme arrives at updated parameters with the learning rate $\alpha$ as follows:
\begin{equation}\label{alreq1}
\theta_t = \theta_{t-1}-\alpha \nabla \mathcal{L}(\theta_{t-1})
\end{equation}
In this study, we focus on determining the optimal learning rates at each step. Note that with the value of $\theta_t$ calculated by Equation (\ref{alreq1}), $\mathcal{L}(\theta_t)$ is the function of $\alpha$, denoted by $g(\alpha)$, that is: 
\begin{equation}\label{alreq2}
g(\alpha)=\mathcal{L}(\theta_{t-1}-\alpha\nabla\mathcal{L}(\theta_{t-1}))
\end{equation}
Then we can approximate $g(\alpha)$ by a quadratic polynomial in $\alpha$ as follows:
\begin{equation}\label{alreq3}
g(\alpha)\approx a_0+a_1\alpha+a_2\alpha^2=g_1(\alpha)
\end{equation}
where $a_0, a_1$, and $a_2$ are characteristic constants at each iteration. The optimal learning rate of $\alpha^*$ is taken such that $g_1(\alpha)$ reaches its minimum as shown in Figure \ref{illustation}. Therefore, if we have the values of theses three characteristic constants, we obtain the following optimal learning rate:
\begin{equation}\label{alreq4}
\alpha^*=-\frac{a_1}{2a_2}
\end{equation}
To determine three parameters $a_0, a_1$, and $a_2$, we need three independent equations, which can be produced from the following conditions:
\begin{equation}\label{alreq5}
g_1(\alpha)\big|_{\alpha=0}=g(0)=\mathcal{L}(\theta_{t-1})=a_0
\end{equation}
\begin{equation}\label{alreq6}
\begin{split}
\frac{\partial g_1(\alpha)}{\partial\alpha}\bigg|_{\alpha=0}&=\frac{\partial g(\alpha)}{\partial\alpha}\bigg|_{\alpha=0}=\frac{\partial\mathcal{L}(\theta_{t-1}-\alpha V)}{\partial\alpha}\bigg|_{\alpha=0}\\
&=-\|\nabla \mathcal{L}(\theta_{ t-1})\|^2=a_1
\end{split}
\end{equation}
For a randomly chosen value of $\bar{\alpha}$, which is called the pre-learning rate in this study, we can have:
\begin{equation}\label{alreq7}
g_1(\alpha)\big|_{\alpha=\bar{\alpha}}=a_0+a_1\bar{\alpha}+a_2\bar{\alpha}^2=g(\bar{\alpha})
\end{equation}
From Equation (\ref{alreq7}), it follows:
\begin{equation}\label{alreq8}
a_2=\frac{g(\bar{\alpha})-a_0-a_1\bar{\alpha}}{\bar{\alpha}^2}
\end{equation}
with the values of $a_0, a_1$, and $a_2$ determined by equations above, now we can have the  optimal learning rate $\alpha^*$ as follows:
\begin{equation}\label{alreq9}
\alpha^*=\frac{\|\nabla \mathcal{L}(\theta_{ t-1})\|^2\bar{\alpha}^2}{2[g(\bar{\alpha})-\mathcal{L}(\theta_{ t-1})+\|\nabla \mathcal{L}(\theta_{ t-1})\|^2\bar{\alpha}]}
\end{equation}

\begin{algorithm}[t]
\caption{QLABGrad}
\label{alg1}
\textbf{Input}: $\mathcal{L}(\theta)$: the loss function; $\theta_0$: initial parameter vector; a pre-learning rate $0<\bar{\alpha}<2/M$ (where $M$ is the Lipschitz smooth constant of $\mathcal{L}(\theta)$). \\
\textbf{Output}: Final weights $\theta_t$

\begin{algorithmic}[1] 
\STATE Initialize $t=0$
\FOR {$t=1, \cdots$ }
\STATE Calculate $\alpha^*$ using Equation (\ref{alreq9})
\STATE $\alpha_t=\alpha^*$
\IF{$\alpha^*<0$}
\STATE $\alpha_t=\bar{\alpha}$
\ENDIF
\STATE Update: $\theta_t=\theta_{t-1}-\alpha_t\nabla \mathcal{L}(\theta_{t-1})$
\ENDFOR
\STATE \textbf{return} the final weights $\theta_t$
\end{algorithmic}
\end{algorithm}

In the gradient descent scheme, the learning rate should be a positive number. However, for a randomly chosen value $\bar{\alpha}$ there is no guarantee that the value of $\alpha^*$ calculated by Equation (\ref{alreq9}) is positive. In the Box-Kanemasu’s method \cite{BeckandArnold1977}, if the value of $\alpha^*$ is negative, the originally chosen value of $\bar{\alpha}$ is halved, and if $\alpha^*$ is still negative, $\bar{\alpha}$ is halved again and again until $\alpha^*$ calculated by Equation (\ref{alreq9}) is positive or stop the algorithm. In the machine learning setting, this strategy could be time-consuming as $g(\bar{\alpha})$ needs to be recalculated for each halving $\bar{\alpha}$. Furthermore, the box Kanemasu's method does not guarantee the convergence. 

\begin{figure*}[htbp]
	\centering
	\includegraphics[width=0.99\textwidth]{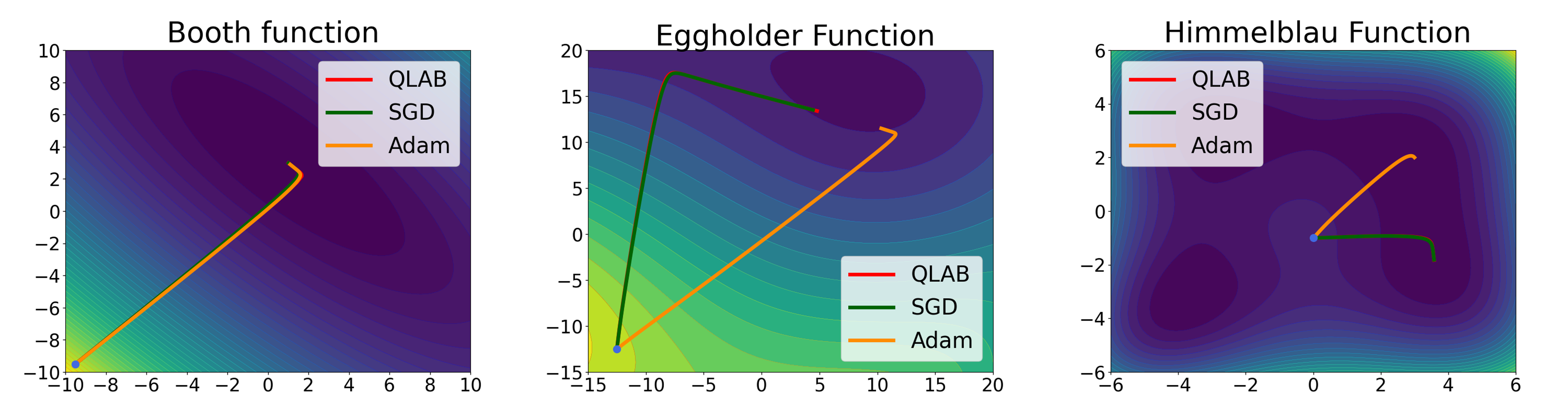}
	\caption{Compares QLABGrad to common optimizers (SGD and Adam) on multimodal functions. The blue dot represents the initial position on the loss surface, and our goal is to illustrate how QLABGrad operates. Compared to other methods, QLABGrad exhibits a more aggressive movement across the objective function surface, which allows it to reach the local minimum value much faster. Specifically, QLABGrad achieves the optimal point on the Booth function, Eggholder function, and Himmelbau function in only 50, 500, and 90 iterations, respectively. In contrast, SGD requires 200, 1500, and 220 iterations, while Adam needs 5000, 10000, and 5000 iterations.}
	\label{trajectory}
\end{figure*}

In QLABGrad, we only calculate an extra value $g(\bar{\alpha})$ at the pre-learning rate $\bar{\alpha}$ once at each iteration step as shown in Algorithm \ref{alg1}. At the first glance, one may see that QLABGrad needs to specify the pre-learning rate $\bar{\alpha}$ with knowing the Lipschitz smooth constant $M$. Actually,  we give a simple Algorithm \ref{alg0} to determine the pre-learning rate $\bar{\alpha}$ from a randomly chosen value without knowing the Lipschitz smooth constant $M$. To get a preliminary understanding of the performance of our proposed QLABGrad in comparison to some standard algorithms such as SGD and Adam, we examine its trajectories on a set of low-dimensional optimization problems as shown in Figure \ref{trajectory}. Remarkably, QLABGrad exhibits faster convergence than both SGD and Adam.

\subsection{Convergence Analysis}

According to the Stone-Weierstrass universal theorem, any continuous function such as any function of deep neural networks (thus any loss function) can be approximated by a smooth function with any accuracy \cite{rudinprinciple}. Therefore, with differentiability only, we cannot derive any reasonable properties of the minimization processes of a loss function. Traditionally, in optimization an extra assumption for the loss function is presented in the form of a Lipschitz condition for its derivative \cite{nesterovintroductory}. $\mathcal{L}(\theta)$ is of $M-$Lipschitz smooth gradient if there exists a positive constant $M>0$ such that:
\begin{equation*}
    \|\nabla\mathcal{L}(\theta_1)-\nabla\mathcal{L}(\theta_2)\|\le M\|\theta_1-\theta_2\|,\quad\forall\theta_1, \theta_2\in\mathbf{R}^d
\end{equation*}
where $\|\cdot\|$ stands for the $\ell_2$ norm of a vector. The minimum constant $M$ that satisfying the above inequality is called the Lipschitz smooth constant of $\mathcal{L}(\theta)$.
\paragraph{Lemma 1:}\cite{Boyd2004}  If $\mathcal{L}(\theta)$ is of $M-$Lipschitz smooth gradient,  then:
\begin{equation}\label{alreq10}
\begin{split}
    \mathcal{L}(\theta_2)&\le \mathcal{L}(\theta_1)+\nabla\mathcal{L}(\theta_1)^T(\theta_2-\theta_1)\\
    &+\frac{M}{2}\|\theta_2-\theta_1\|^2, \quad \forall\theta_1,\theta_2\in\mathbf{R}^d
\end{split}
\end{equation}

Let $\theta_1=\theta_{t-1}$ and $\theta_2=\theta_{t-1}-\alpha\nabla \mathcal{L}(\theta_{ t-1})$. As $\mathcal{L}(\theta)$ is of $M-$Lipschitz smooth gradient, for any $\alpha>0$ Equation (\ref{alreq10}) becomes:
\begin{equation}\label{alreq11}
\begin{split}
g(\alpha)&=\mathcal{L}(\theta_{t-1}-\alpha\nabla \mathcal{L}(\theta_{ t-1})) \\
&\le \mathcal{L}(\theta_{t-1})-\alpha \|\nabla \mathcal{L}(\theta_{ t-1})\|^2\\
&\qquad+\frac{\alpha^2M}{2}\|\nabla \mathcal{L}(\theta_{ t-1})\|^2\\
&= g(0)-\alpha\left(1- \frac{M\alpha}{2}\right)\|\nabla \mathcal{L}(\theta_{ t-1})\|^2
\end{split}
\end{equation}

\paragraph{Lemma 2:}  Assume that $\mathcal{L}(\theta)$ is of $M-$Lipschitz smooth gradient. If $a_2>0$, then  $1/M\le\alpha^*\le\bar{\alpha}$.

\emph{Proof:} 
As $a_2>0$ and thus the value of $\alpha^*$ calculated by Equation (\ref{alreq9}) is positive, combining Equations (\ref{alreq9}) and (\ref{alreq11}) yields:
\begin{equation}\label{alreq12}
\begin{split}
\alpha^*&=\frac{\|\nabla \mathcal{L}(\theta_{t-1})\|^2\bar{\alpha}^2}{2[g(\bar{\alpha})-\mathcal{L}(\theta_{t-1})+\|\nabla \mathcal{L}(\theta_{ t-1})\|^2\bar{\alpha}]}\\
&\ge \frac{\|\nabla \mathcal{L}(\theta_{t-1})\|^2\bar{\alpha}^2}{2\frac{\bar{\alpha}^2M}{2}\|\nabla \mathcal{L}(\theta_{t-1})\|^2}=\frac{1}{M}
\end{split}
\end{equation}
As $\alpha^*$ is the minimizer of $g_1(\alpha)$ when $a_2>0$, we have:
\begin{equation}\label{alreq13}
\begin{split}
    0&\le g_1(\bar{\alpha})-g_1(\alpha^*)\\
    &=a_0+a_1\bar{\alpha}+a_2\bar{\alpha}^2-(a_0+a_1\alpha^*+a_2\alpha^{*2})\\
    &=a_1(\bar{\alpha}-\alpha^*)+a_2(\bar{\alpha}^2-\alpha^{*2})\\
    &=(\bar{\alpha}-\alpha^*)(a_1+a_2(\bar{\alpha}+\alpha^*))\\
    &=a_2(\bar{\alpha}-\alpha^*)(-\frac{1}{2}\alpha^*+(\bar{\alpha}+\alpha^*))\\
    &=a_2(\bar{\alpha}-\alpha^*)(\frac{1}{2}\alpha^*+\bar{\alpha})
\end{split}
\end{equation}
As $a_2>0$, and $\frac{1}{2}\alpha^*+\bar{\alpha}\ge \frac{1}{2M}+\bar{\alpha}>0$. From Equation (\ref{alreq13}), we conclude that $1/M\le\alpha^*\le\bar{\alpha}$ while combining with Equation (\ref{alreq12}).

\paragraph{Theorem 1:}
\label{thm1}
Assume that $\mathcal{L}(\theta)$ is of $M-$Lipschitz smooth gradient. If the pre-learning rate $0<\bar{\alpha}<2/M$ holds, QLABGrad guarantees that: 
\begin{equation}\label{alreq14}
    \min_{0\le t\le T-1}\|\nabla\mathcal{L}(\theta_{t})\|\le\frac{1}{\sqrt{T}}\sqrt{\frac{\mathcal{L}(\theta_0)-\mathcal{L}^*}{C} }
\end{equation}
where $\mathcal{L}^*=min_{\theta\in\mathbf{R}^d}\mathcal{L}(\theta)$, $C=\min\{\bar{\alpha},\frac{2-M\bar{\alpha}}{2M}\}$, and $T$ is the total number of iterations.

\emph{Proof:} 
In case that the value of $\alpha^*$ calculated by Equation (\ref{alreq9}) is negative, we have: 
\begin{equation*}
    g(\alpha_t)-\mathcal{L}(\theta_{ t-1})+\|\nabla \mathcal{L}(\theta_{ t-1})\|^2\bar{\alpha}<0
\end{equation*}
which indicates:
\begin{equation}\label{alreq15}
   \bar{\alpha} \|\nabla \mathcal{L}(\theta_{ t-1})\|^2<\mathcal{L}(\theta_{ t-1})-\mathcal{L}(\theta_t)
\end{equation}
as $g(\alpha_t)=g(\bar{\alpha})=\mathcal{L}(\theta_t)$. 

In case that the value of $\alpha^*$ calculated by Equation (\ref{alreq9}) is positive, Equation (\ref{alreq11}) with $\alpha=\alpha^*$ becomes:
\begin{equation}\label{alreq16}
\begin{split}
\mathcal{L}(\theta_{t-1})-\mathcal{L}(\theta_{t})&\ge \alpha^*(1-\frac{M\alpha^*}{2}) \|\nabla \mathcal{L}(\theta_{ t-1})\|^2\\
&\ge\frac{2-M\bar{\alpha}}{2M}\|\nabla \mathcal{L}(\theta_{ t-1})\|^2
\end{split}
\end{equation}

Combining Equations (\ref{alreq15}) and (\ref{alreq16}), for any cases, we have:
\begin{equation}\label{alreq17}
   C\|\nabla \mathcal{L}(\theta_{ t-1})\|^2\le\mathcal{L}(\theta_{ t-1})-\mathcal{L}(\theta_t)
\end{equation}
where $C=\min\{\bar{\alpha},\frac{2-M\bar{\alpha}}{2M}\}$.

Summing inequalities (\ref{alreq17}) for $t=0,\cdots T-1$ yields:
\begin{equation}\label{alreq18}
    C\sum^{T-1}_{t=0}\|\nabla \mathcal{L}(\theta_{ t-1})\|^2\le \mathcal{L}(\theta_0)-\mathcal{L}(\theta_T)\le \mathcal{L}(\theta_0)-\mathcal{L}^*
\end{equation}
which concludes the proof.

\subsection{Determination of the Pre-learning Rate}
As shown in the proof of Theorem 1, the pre-learning rate should be bounded by $2/M$ to guarantee the convergence of Algorithm \ref{alg1}, where $M$ is the Lipschitz smooth constant of the loss function. Although for a given function, its Lipschitz smooth constant can be accurately estimated by its second derivatives, it is not easy to accurately estimate the Lipschitz smooth constant of an oracle function such as the loss functions of deep learning models. Nevertheless, Algorithm \ref{alg1} only needs to assure that the pre-learning rate is bounded by $2/M$ instead of the exact value of $M$.

\begin{algorithm}[t]
\caption{FindPLR ($\mathcal{L}({\theta}), \alpha_0$)}
\label{alg0}
\textbf{Input}: $\mathcal{L}(\theta)$: the loss function; $\alpha_0$: an initial pre-learning rate ($0.1, 0.01, 0.005$, or the like). \\
\textbf{Output}: Pre-learning rate $\bar{\alpha}$

\begin{algorithmic}[1] 
\STATE Randomly choose a parameter value $\theta$ and let $\bar{\alpha}=\alpha_0$
\STATE Calculate $ g(0)=\mathcal{L}(\theta)$ and $\nabla \mathcal{L}(\theta)$
\STATE Calculate $g(\bar{\alpha})=\mathcal{L}(\theta-\bar{\alpha}\nabla \mathcal{L}(\theta))$
\WHILE{$a_2=g(\bar{\alpha})-g(0)+\bar{\alpha}\|\nabla \mathcal{L}(\theta)\|^2<0$}
\STATE $\bar{\alpha}=2\bar{\alpha}$
\STATE Calculate $g(\bar{\alpha})=\mathcal{L}(\theta-\bar{\alpha}\nabla \mathcal{L}(\theta))$
\ENDWHILE
\WHILE{$g(\bar{\alpha})>g(0)$}
\STATE $\bar{\alpha}=0.5\bar{\alpha}$
\STATE Calculate $g(\bar{\alpha})=\mathcal{L}(\theta-\bar{\alpha}\nabla \mathcal{L}(\theta))$
\ENDWHILE
\end{algorithmic}
\end{algorithm}

On the one hand, when the pre-learning is too large (greater than $2/M$), the convergence is not guaranteed. On the other hand, when the pre-learning is too small, by Lemma 2 the learning rate $\alpha^*$ is smaller, which leads to the slow convergence. Therefore we propose Algorithm \ref{alg0} to determine the proper pre-learning rate, called FindPLR, which is based on the following Lemma 3.

\paragraph{Lemma 3:}  Assume that $\mathcal{L}(\theta)$ is of $M-$Lipschitz smooth gradient. For any $\alpha>0$, if $g(\alpha)>g(0)$, then $\alpha\ge 2/M$.

\emph{Proof:} 
If $\alpha<2/M$, then $\alpha\left(1- \frac{M\alpha}{2}\right)>0$. From Equation (\ref{alreq11}) it gives $g(\alpha)<g(0)$, which contradicts $g(\alpha)>g(0)$.

For a given loss function $\mathcal{L}(\theta)$ with a parameter vector $\theta$, starting with a randomly chosen initial pre-learning rate, FindPLR finds a proper pre-learning rate with two `while' loops. The first loop is to assure that the final pre-learning is not too small, while the second loop is to assure that the final learning rate is bounded by $2/M$ according to Lemma 3. Note that 10 iterations of either loop changes the the initial pre-learning rate more than $1000$ ($\approx 2^{10}$) times. Therefore, if one does not purposely make thing difficult by choosing an extremely large or extremely small initial pre-learning rate, FindPLR can find a proper pre-learning rate in about 10 iterations, where each iteration only needs to calculate one extra value of the loss function. In practice FindPLR mainly runs either loop, specifically, if the initial pre-learning rate is too small, the first loop runs for a few iterations while the second loop runs at most once, and otherwise the second loop runs a few iterations.

\subsection{QLABGrad for Deep Learning with Minibatch}
For the batch training, the expression of the loss function of $\mathcal{L}(\theta)$ does not change with epochs, therefore it is enough to call Algorithm \ref{alg0} once at the beginning of Algorithm \ref{alg1}. However, for the minibatch training, the expression of the loss function of $\mathcal{L}(\theta)$ depends on the samples in a minibatch although it is expected that the loss function of $\mathcal{L}(\theta)$ is not changed very much among the different minibatches. Nevertheless the different proper pre-learning rate can easily be determined for each minibatch with Algorithm \ref{alg0} in a few steps. In practice, in the whole training process  it is recommended that Algorithm \ref{alg0} should be called a number of times to assure that the pre-learning rate is proper.

\section{Experiments and Results}
To assess the effectiveness of our proposed QLABGrad, we carry out extensive experiments on three different datasets (MNIST \cite{mnist}, CIFAR10 \cite{cifar10}, and Tiny-ImageNet \cite{tiny}) with various models (multi-layer neural network, CNN, and ResNet-18), comparing to various popular competing schemes, including basic SGD, RMSProp, AdaGrad, and Adam.  To ensure that these competing schemes demonstrate their own optimal loss reduction effects, we replicate the experiments with various settings of their hyperparameters.

\begin{figure*}[htbp]
	\centering
	\includegraphics[width=0.99\textwidth]{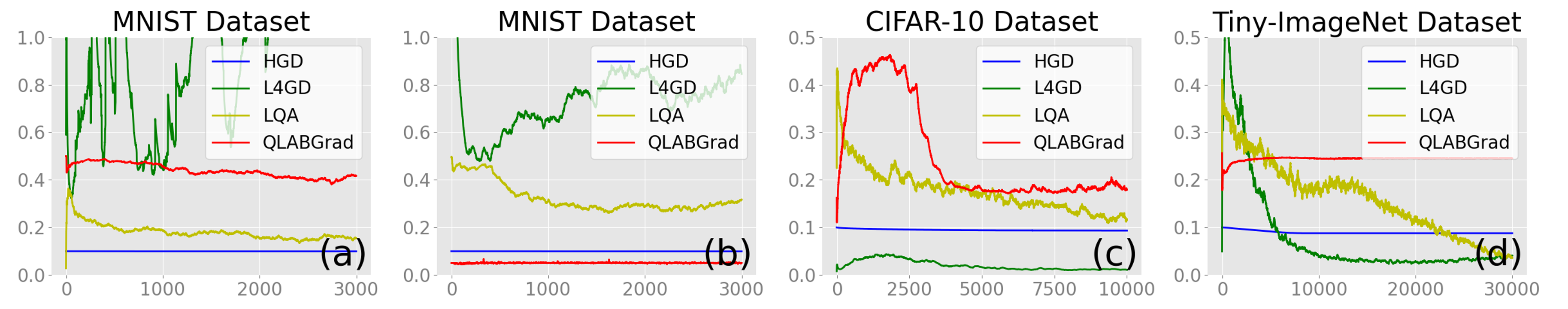}
	\caption{Learning rate variations of HGD, L4GD, LQA and QLABGrad for training MLP (a) and CNN (b) model on MNIST dataset and training ResNet18 on CIFAR-10 (c) and Tiny-ImageNet dataset (d). The X-axis represents the number of iterations and the Y-axis indicates the learning rate values.}
	\label{lr}
\end{figure*}

\begin{table*}[htbp]
\centering
\begin{tabular}{lcccccccc}
\hline 
Model & SGD & RMSProp & Adagrad & Adam & HGD & L4GD & LQA & QLABGrad \\
\hline 
MLP   & 98.42 & 96.01 & 98.53 & 95.88 & 98.43 & 97.43  & 98.47 & 98.36 \\
CNN   & 99.09 & 97.44 & 98.86 & 97.51 & 99.06 & 98.54  & 98.05 & 99.18 \\
\hline
\end{tabular}
\caption{Comparison of test accuracy for various optimizers on the MNIST dataset.}
\label{test_acc}
\end{table*}

\begin{figure*}[t]
	\centering
	\includegraphics[width=0.99\textwidth]{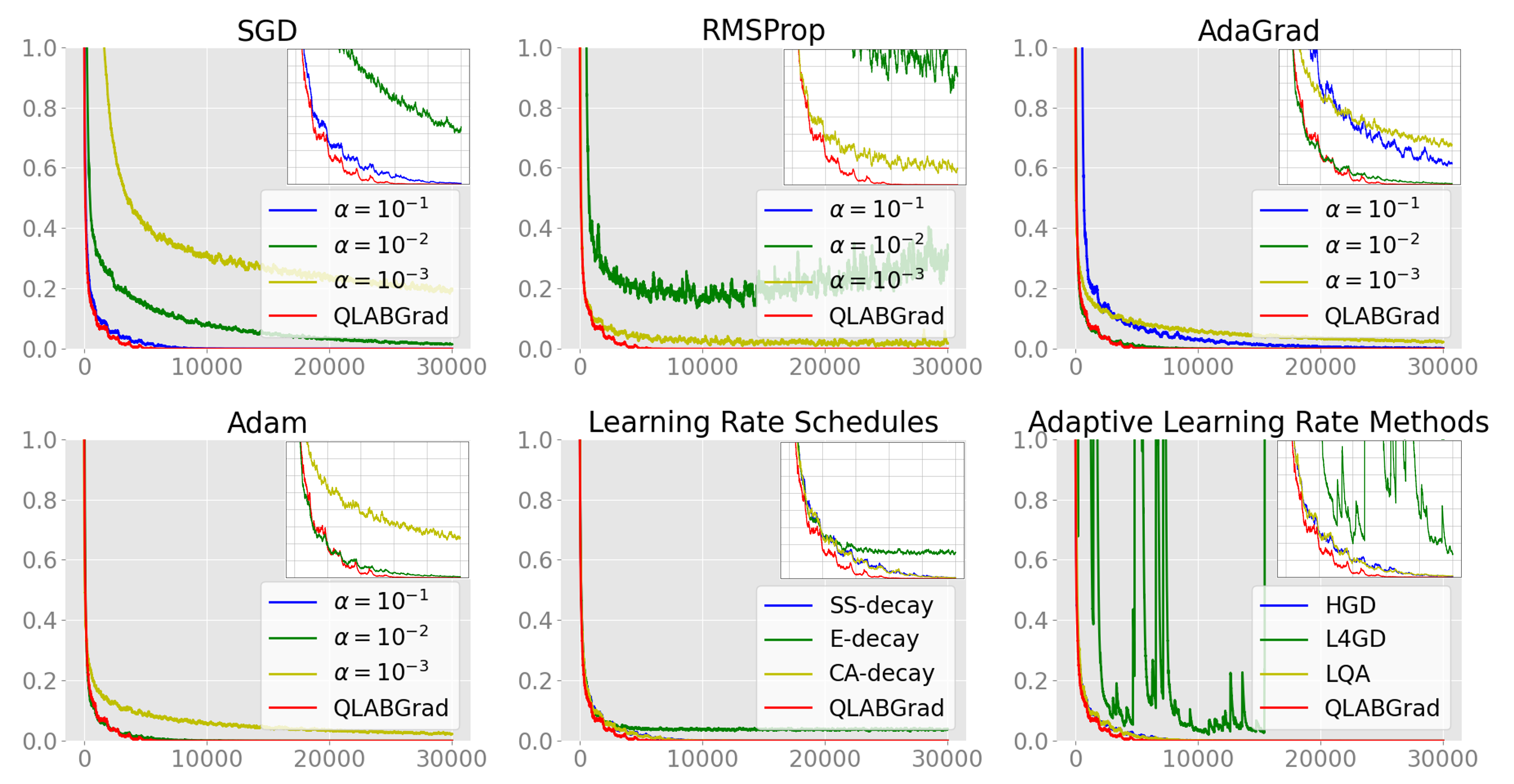}
	\caption{Training loss for MLP on the MNIST dataset. The X-axis represents the number of iterations and the Y-axis indicates the learning rate values. A zoomed-in view of the loss changes during the initial 10,000 iterations is demonstrated in the top-right corner of each corresponding subplot, with the loss variation ranging from 0 to 0.2.}
	\label{mlp_loss}
\end{figure*}

Furthermore, we examine some recently developed schemes such as HGD, L4GD, and LQA.  To make sure fair comparisons, we set the initial learning rate to $\alpha=10^{-1}$ for all competing methods. In order to provide a clearer illustration of the variations in learning rates among these schemes for different models and datasets, we present the learning rates of these methods in Figure \ref{lr}.

Additionally, we also compare QLABGrad with decay schemes such as SS-decay, E-decay, and CA-decay, all implemented based on the basic SGD method. From our experiments, SGD with an initial learning rate of $10^{-1}$ exhibited the best performance. Therefore, we adopted $10^{-1}$ as the learning rate ($\alpha$ in Table \ref{Tablem}) for these schemes. Specifically, SS-decay halves the learning rate every $T$ iterations, E-decay decreases the learning rate by a factor of $\beta=\frac{1}{2}$, and CA-decay sets the minimum learning rate $\beta=10^{-3}$. All experiments are performed on a single RTX 3090 GPU.

\subsection{Evaluation on MNIST Digit Recognition}
\begin{figure}[t]
	\centering
	\includegraphics[width=0.99\linewidth]{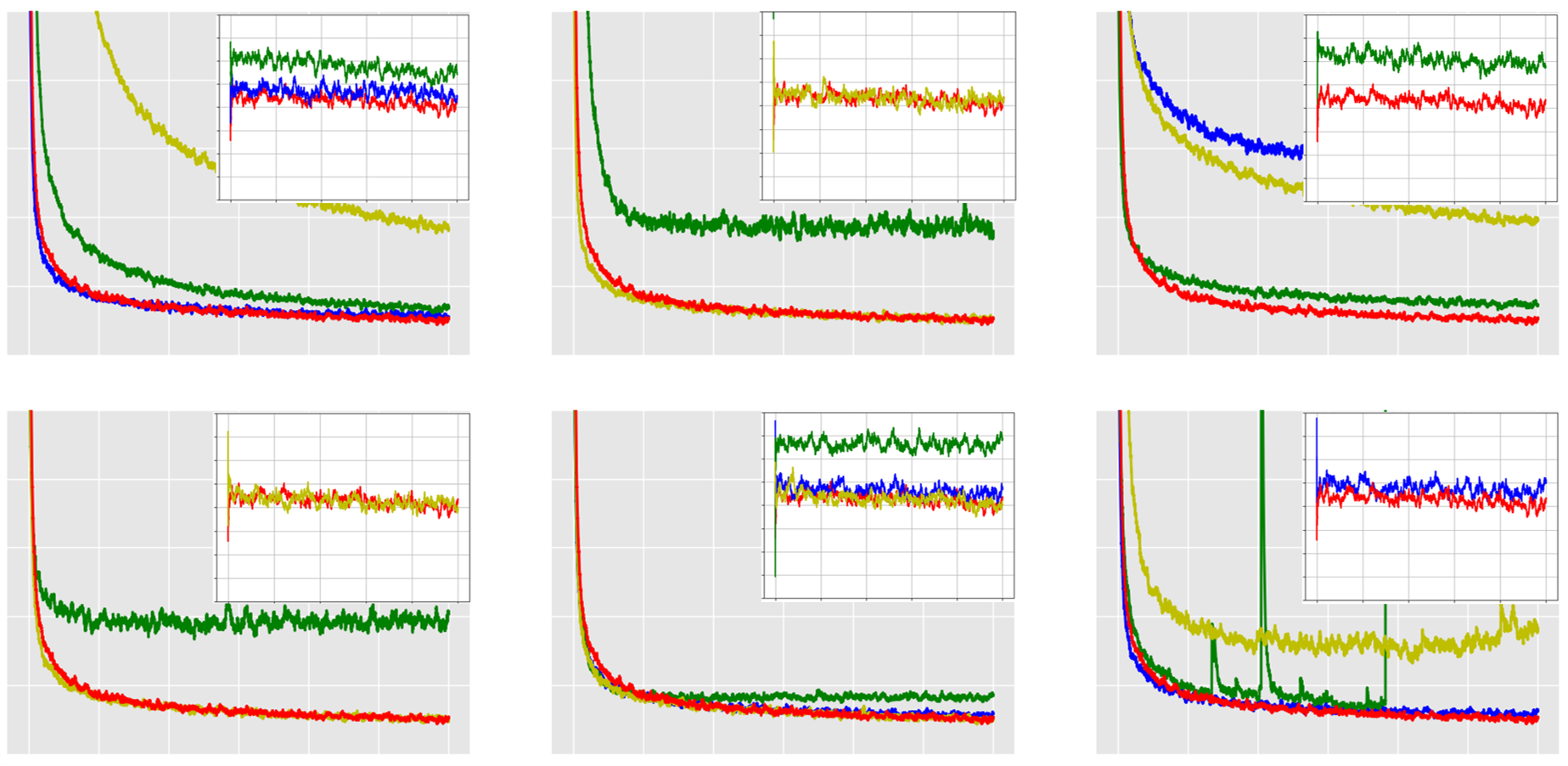}
	\caption{Training loss for CNN on the MNIST dataset,  which adheres to the settings established in Figure 4. The overall loss variations in each subplot range from 0 to 1, with iterations spanning from 0 to 30,000. The zoomed-in view shows the loss changes across the last 10,000 iterations, with the loss variation ranging from 0 to 0.2.}
	\label{cnn_loss}
\end{figure}

The MNIST dataset \cite{mnist} includes 60,000 training and 10,000 testing images. Figures \ref{mlp_loss} and \ref{cnn_loss} show the training loss for a multi-layer neural network (MLP) and a convolution neural network (CNN), respectively, with a mini-batch size of 64. The loss is given by cross-entropy. Notably, when the MLP and CNN models employ RMSProp or Adam optimizers with an initial learning rate of $10^{-1}$, the loss curves become out of range and are not displayed in the graphs. Although we are not primarily concerned about model generalization capability by adopting different schemes in this work, we still report the test accuracy among various methods on dataset MNIST in Table \ref{test_acc}, showing comparable performance across schemes. Further details and discussions on network structures follow.

\paragraph{Multi-Layer Neural Network:} The architecture of the MLP model is built with two fully connected hidden layers, each of which contains 1000 neurons, and the ReLU function is adopted as the activation function.

QLABGrad demonstrates superior performance compared to the standard stochastic gradient descent (SGD) schemes and SGD with different weight decay schemes, highlighting the effectiveness of the proposed QLABGrad scheme. In contrast, among the popular schemes, RMSProp exhibits the poorest performance primarily due to its susceptibility to local optima. RMSProp faces challenges in escaping sharp minima \cite{rms_local}, which are narrow regions within the optimization landscape. This difficulty stems from its adaptive learning rate mechanism, where the learning rate for each parameter is adjusted based on the average of squared gradients. Consequently, this scaling factor proves problematic when confronted with the sharp minima, as the gradients in such regions tend to be small. RMSProp becomes trapped in a local optimum along the MLP loss curve. Conversely, Adagrad performs admirably, owing to its ability to handle sparse data \cite{adagrad_sparse}. The MNIST dataset comprises relatively sparse grayscale images, containing numerous zero-valued pixels (image background). Adam, built upon RMSProp, introduces a momentum mechanism that mitigates the issue of becoming trapped in local optima and achieves competitive performance in this particular scenario.

Among various adaptive learning rate schemes, L4GD shows relatively poor results due to its linear approximation of the loss function. Despite a conservative hyperparameter value of 0.15, its aggressive learning rate causes drastic loss oscillations. As Figure \ref{lr} (a) shows, L4GD's learning rate varies significantly, while the HGD method shows minimal variation, largely due to the hyperparameter $\beta$, set to $10^{-3}$, which smooths learning rate fluctuations within a $10^{-2}$ range. LQA and QLABGrad display similar trends, with learning rates initially increasing in early iterations before settling into a reasonable range with minor variations, a result of substantial initial loss decay.

\paragraph{Convolution Neural Network:} CNN structure uses two 2D convolution layers. The kernel size of the convolution operation is 5 and the stride is 1. The average pooling operation is followed by each layer, the extracted feature maps use fully connected layers and the Relu activation function for image classifications. 

Compared to the multilayer perceptron (MLP) model, convolutional neural networks (CNNs) employ a shared-weights convolution kernel and hierarchical feature extraction mechanisms that significantly reduce feature sparsity. Consequently, Adagrad  yields a relatively poor loss reduction. Moreover, the smoother loss curve in CNNs leads to a negligible $a_1$ term in the QLABGrad method. As shown in Figure \ref{lr} (b), QLABGrad has lower learning rates. However, QLABGrad demonstrates better convergence and smaller converged loss values. LQA is impacted by larger learning rates, as depicted in Figure \ref{lr} (b), thus affecting the model's convergence. Table \ref{test_acc} also indicates that QLABGrad achieves the best test results.

\begin{figure*}[htbp]
	\centering
	\includegraphics[width=0.99\textwidth]{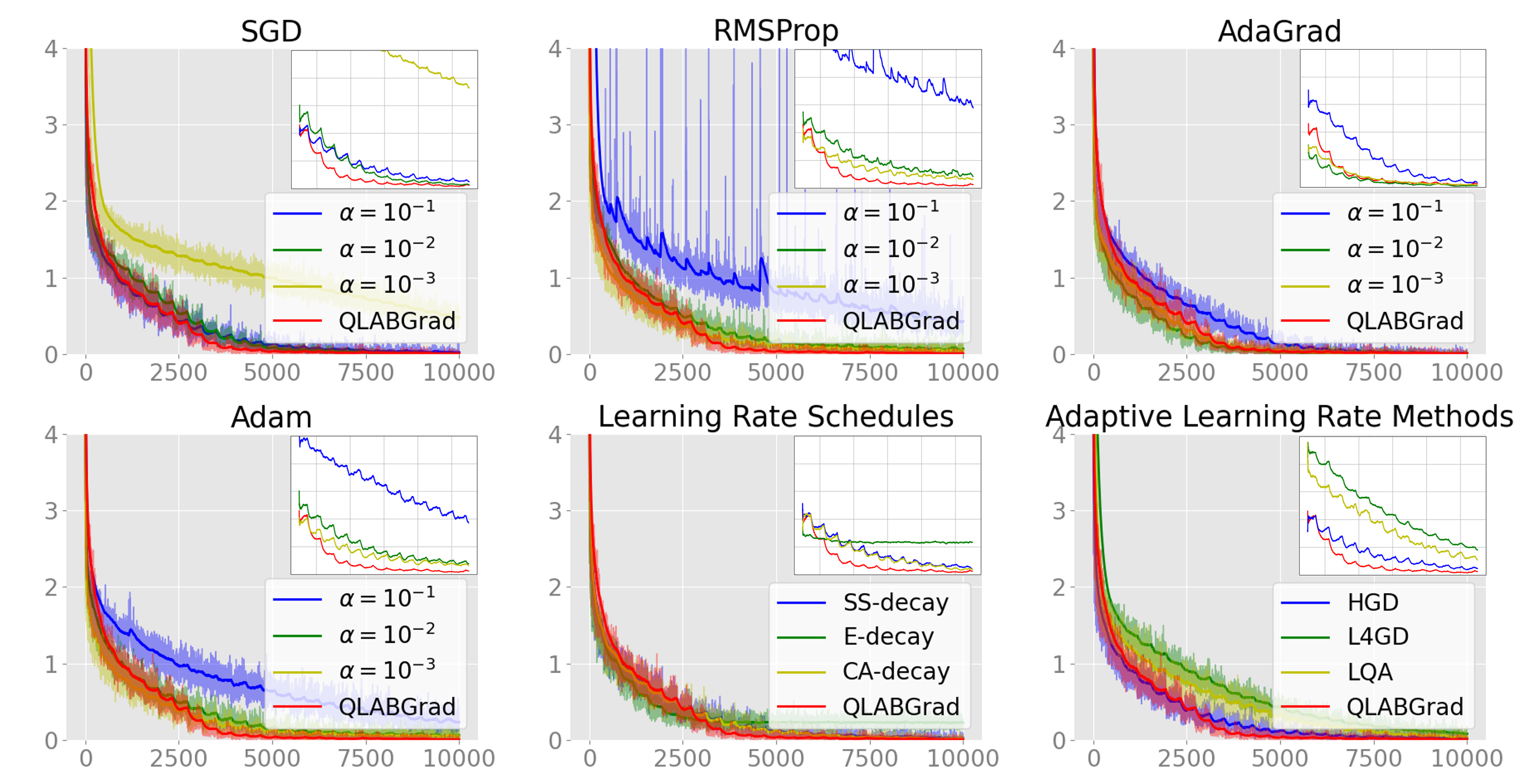}
	\caption{Training loss for ResNet18 on the CIFAR-10 dataset. The X-axis represents the number of iterations and the Y-axis indicates the learning rate values. A zoomed-in view highlights the loss changes between the 2,000th and 8,000th iterations, with the loss variation ranging from 0 to 1.}
	\label{cifar10_loss}
\end{figure*}

\subsection{Evaluation on CIFAR-10}
The CIFAR-10 dataset \cite{cifar10} comprises 50,000 training images and 10,000 test images, each $32 \times 32$ pixels with RGB colors, distributed across 10 classes. We employ ResNet-18 \cite{resnet}, a renowned architecture for training over 10,000 iterations, using a batch size of 128. Training settings are consistent with those used for the MNIST dataset training.

 Through our analysis of Figure \ref{cifar10_loss}, it is evident that QLABGrad exhibits a remarkable advantage in reducing loss compared to all other schemes. Adagrad also demonstrates strong performance; however, it heavily depends on selecting an appropriate initial learning rate. Notably, using $10^{-1}$ as the initial learning rate diminishes the effectiveness of Adagrad. On the other hand, the remaining schemes display a noticeable disparity, compared to QLABGrad, which further confirms the outstanding performance of QLABGrad.

In Figure \ref{lr} (c), QLABGrad's learning rate initially rises then falls, reflecting the model's early-stage complex loss surface. Initially large gradients increase the learning rate, as indicated by the $a_1$ term in Equation (5). However, ResNet's residual connections, preventing gradient vanishing, eventually stabilize QLABGrad's rate. This dynamic adjustment of the learning rate is essential for optimal training progress and model accuracy. Similar trends in learning rate variations are also evident in L4GD and LQA, but contrast with the HGD method, which shows consistent learning rates across steps, potentially affecting its adaptability to varying training scenarios.

\subsection{Evaluation on Tiny-ImageNet} The Tiny-ImageNet dataset \cite{tiny} comprises 200 categories, with 500 images per category. Each image in the dataset has dimensions of $64 \times 64$ pixels. During the training process, each model underwent 30,000 iterations, utilizing a batch size of 128. All other settings remained consistent with those employed for training on  MNIST \cite{mnist}.

\begin{figure}[htbp]
	\centering
	\includegraphics[width=0.99\linewidth]{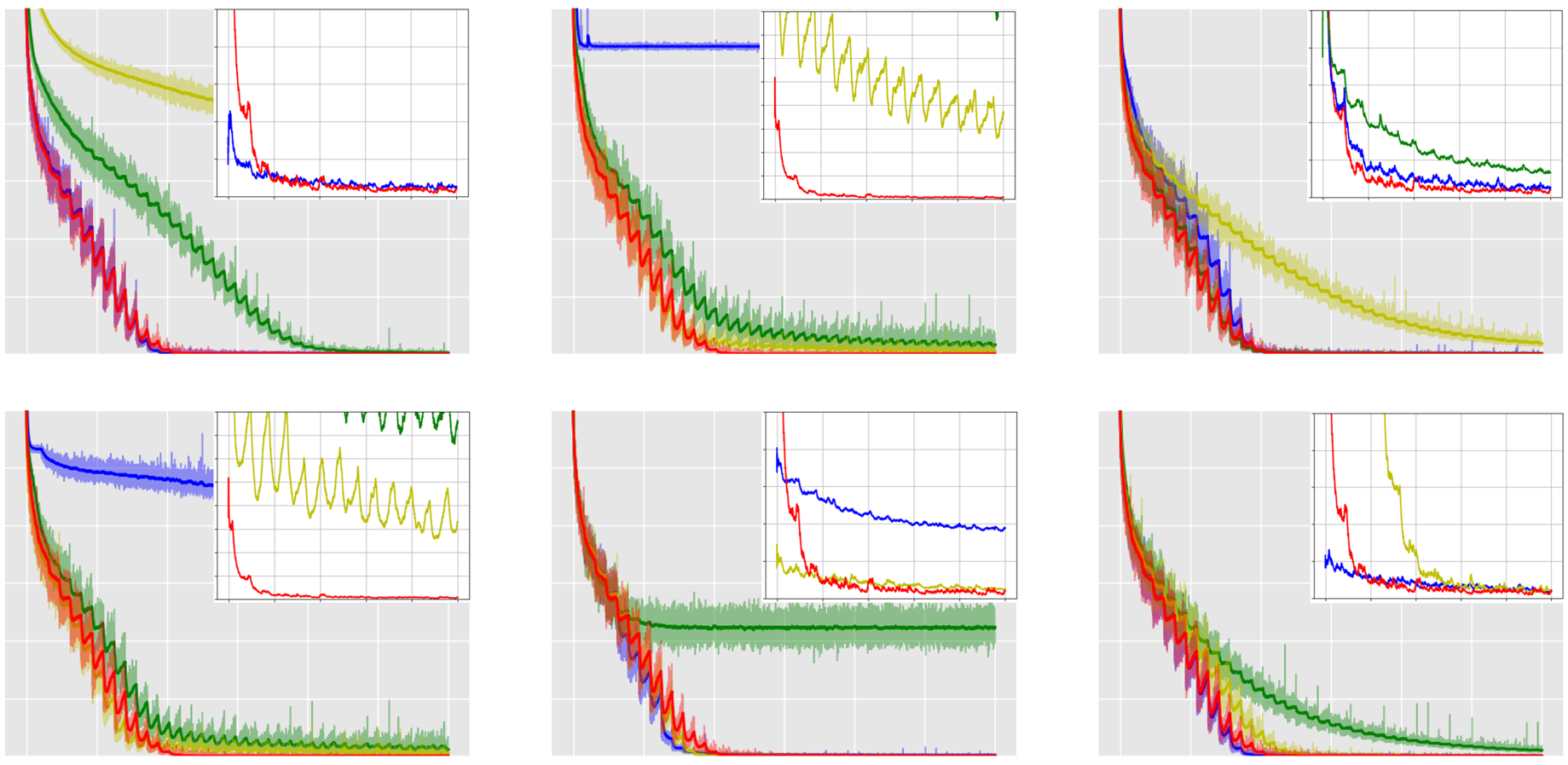}
	\caption{Training loss for ResNet18 on the Tiny-ImageNet dataset is depicted in each subplot, which adheres to the settings established in Figure 6. The overall loss variations for each subplot range from 0 to 6, and the iterations from 0 to 30,000. A detailed view of the initial 20,000 iterations highlights finer loss changes within the range of 0 to 0.05.}
	\label{loss_imagenet}
\end{figure}

 \begin{figure*}[t]
	\centering
	\includegraphics[width=0.99\textwidth]{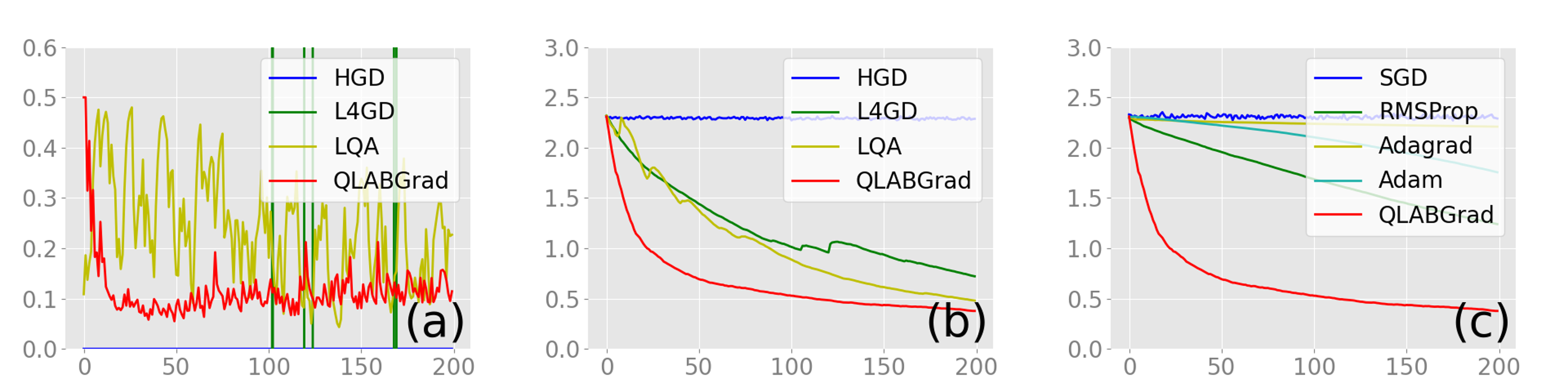}
	\caption{Learning rate variations (a) range from 0 to 0.6, and training loss (b and c) range from 0 to 3 for the MLP model with different optimizers on the MNIST dataset during the initial 200 iterations. By using Algorithm \ref{alg0}, our proposed QLABGrad boosts the initial learning rate from $10^{-5}$ to 0.00128 after 7 iterations. Combined with Algorithm \ref{alg1}, it efficiently adjusts the learning rate into a reasonable range, yielding favorable optimization outcomes in the early stage.}
	\label{plr}
\end{figure*}

Figure \ref{loss_imagenet} shows that at around 10,000 steps, QLABGrad significantly reduces the loss function, outperforming other methods. While SGD effectively reduces loss, it relies on a fixed learning rate. RMSProp, Adagrad, and Adam, using historical gradients, converge more slowly due to different optimization paths.

Adaptive learning rate methods, as seen in Figure \ref{lr} (d), show L4GD's aggressive strategy leads to a sharp learning rate increase in the initial 2000 steps, risking overshooting the minimum and slower loss reduction. Similarly, LQA's initial high learning rate slows convergence. In contrast, QLABGrad finds a more suitable learning rate, enhancing loss reduction and convergence.

\subsection{Combining QLABGrad with FindPLR}
Figure \ref{plr} shows how combining Algorithm \ref{alg0} with Algorithm \ref{alg1} adjusts learning rates, particularly when facing unfavorable pre-learning rates. QLABGrad with FindPLR efficiently iterates to keep initial values within an optimal range for specific tasks. While standard SGD is effective in deep learning, it depends heavily on choosing the right learning rate, often determined through time-consuming grid searches or trial-and-error, requiring deep expertise. The quick identification of a suitable learning rate, as demonstrated by QLABGrad with FindPLR, is a key advantage. Figure \ref{plr} also shows a stable loss reduction, indicating QLABGrad's superior adjustment and optimization performance compared to common (SGD, RMSProp, Adagrad, Adam) and recent (HGD, L4GD, LQA) methods, even with suboptimal pre-learning rates.

\subsection{Time Overhead Analysis}
In assessing time efficiency across MNIST, CIFAR-10, and Tiny-ImageNet datasets, we used SGD as the benchmark for training duration. RMSProp, Adagrad, and Adam demonstrated comparable time costs to SGD. HGD and L4 took about 1.3 times longer, while QLABGrad required approximately 1.4 times more time. Notably, LQA showed the most significant increase in training duration, taking about 1.6 times longer than SGD.

\section{Discussions} The common schemes for deep learning, including SGD, RMSProp \cite{rms}, Adagrad \cite{adagrad}, and Adam \cite{adam}, are highly sensitive to the selection of hyperparmeters, which can have a significant impact on the learning outcomes. Although the learning rate schemes can improve the reduction of loss functions to some extent, they still heavily depend on the choice of their more hyperparameters. Additionally, these schemes lack consistency across different models and datasets. For example, Adagrad performs well on datasets with sparse data features like MNIST but exhibits slower convergence on the Tiny-ImageNet dataset. RMSProp and Adam, which incorporate the historical gradients as references, demonstrate good convergence speed on CIFAR10 while experiencing performance degradation on Tiny-ImageNet. Therefore, the use of these schemes requires not only a thorough understanding of the dataset and model but also empirical exploration or extensive try-and-error experiments to determine the critical hyperparameters.

HGD \cite{hgd} is a modification of the Adagrad \cite{adagrad}, where the current gradient inner product is replaced with the gradient inner product of the last two updates. The performance of HDG is expected to be similar to that of Adagrad. However, due to the introduction of the additional hyperparameter $\beta$, the change in the learning rate is in a narrow range, resulting in only limited performance improvements over SGD. L4GD \cite{l4} is designed based on the linear approximation of the loss function, and its update strategy is more aggressive. Although it introduces a hyperparameter to smooth the value of the learning rate, it can still lead to gradient explosions in cases where the loss surface is relatively smooth on datasets such as MNIST.

LQA \cite{lqa} determines the learning rate based on local quadratic approximation of the loss function. It utilizes three loss values: the current loss value, the loss value after a gradient ascent at the current point, and the loss value after a gradient descent at the current point, to obtain a quadratic fitting curve. However, this approach necessitates the calculation of three loss values at each step and does not effectively utilize the gradient information. Consequently, LQA is highly insensitive to changes in gradients. In contrast, our proposed QLABGrad thoroughly optimizes the quadratic function fitting by incorporating an additional forward calculation with the gradient information. This not only reduces redundant computations but also effectively integrates gradient information. In terms of experimental results, QLABGrad exhibits excellent consistency and ensures convergence for both simple and complex loss surfaces as expected by our theoretical analysis.

\section{Conclusion}
We proposed QLABGrad, a novel quadratic loss approximation-based learning rate scheme, offering better or comparable performance in deep learning without extra hyperparameters and ensuring convergence under smooth M-Lipschitz conditions. The key insight is to determine the optimal learning rate for each parameter update step by incorporating both the loss value and gradient information. Our experiments across various architectures with three classical datasets demonstrate that QLABGrad generally outperforms other methods, including those using learning rate decay, adaptive learning rates, and constant learning rates with adapted descent directions in deep learning. It also exhibits superior effectiveness in the learning rate determination and performance consistency on both simple and complex loss surfaces. The ease of implementation of QLABGrad makes it an attractive method with the potential to become a standard tool for training deep learning models.

\section{Acknowledgments}
This study is supported by Natural Science and Engineering Reseach Council of Canada (NSERC).

\bibliography{aaai24}
\end{document}